# Sign in the Air to Unlock: An Interface for authentication in Virtual and Augmented Reality Powered by Point–Voxel Cross-Attention Network

NEDA ABDOLRAHIMI*, Syracuse University, USA
THIRU SIDDHARTH, Syracuse University, USA
FRANK SICONG CHEN, Syracuse University, USA
VIR V PHOHA, Syracuse University, USA

Significant advancement of immersive technologies such as Virtual and Augmented Reality (VR/AR) and their integration into diverse aspects of modern life need authentication interfaces that are secure, intuitive, and compatible with embodied interaction. Traditional methods such as passwords, PINs, and device-based logins, break immersion and rely on external hardware. Recent 3D-specific behavioral approaches, such as hand-gesture, eye-tracking, and electroencephalography (EEG)-based methods, offer promising alternatives but often require specialized sensors or constrain natural movement, limiting usability in dynamic environments. We present Sign in the Air to Unlock, an in-air signature interface that enables users to authenticate by signing naturally in 3D space which is a familiar, personal, and reproducible gesture. To realize this interface, we design a point–voxel Cross-Attention Network (PV-Net) that jointly models local motion dynamics and global spatial structure from 3D trajectories. The model is evaluated on two datasets: the public DeepAirSig dataset (1,800 signatures from 40 users) and ImmAirsig, a new dataset collected using Meta Quest 2 in immersive VR (880 samples from 22 users). PV-Net achieves an Equal Error Rate of 2.5% on DeepAirSig and 76% classification accuracy on ImmAirSig. These findings highlight the potential of 3D behavioral interfaces for seamless, user-centric authentication that merges security with natural interaction in immersive environments.

Authors' addresses: Neda Abdolrahimi, nabdolra@syr.edu, Syracuse University, 900 S Crouse Ave, Syracuse, New York, USA, 13244; Thiru Siddharth, tsiddhar@syr.edu, Syracuse University, 900 S Crouse Ave, Syracuse, New York, USA, 13244; Frank Sicong Chen, schen154@syr.edu, Syracuse University, 900 S Crouse Ave, Syracuse, New York, USA, 13244; Vir V Phoha, vvphoha@syr.edu, Syracuse University, 900 S Crouse Ave, Syracuse, New York, USA, 13244.

## 1 INTRODUCTION

The rapid evolution of VR/AR technologies has allowed immersive platforms to become more integrated into many aspects of life including entertainment, gaming, education [1, 3, 6, 19, 35], tourism [7, 23, 36], advertising [2, 22], healthcare [11, 32, 33], in some cases even at surgical operating tables [30]. Unlike personal smartphones or smartwatches, VR/AR headsets are frequently shared between multiple users like family members, co-workers, or students. This shared usage brings up a pressing need for a secure user interface for authentication to protect personal data and maintain user-specific settings during an immersive session. Despite advances in immersive interaction, designing intuitive and frictionless user interfaces for authentication remains an ongoing challenge.

Many commercial VR/AR devices rely on paired external devices such as smartphones for authentication before launching the immersive session [34]. While secure, this approach disrupts the user experience, requiring users to remove their headsets to interact with the secondary device. Within the VR environment itself, some systems adapt traditional mechanisms like password entry [39] or pattern locks [40] to virtual interfaces. Although these methods are as secure as authentication on devices with touchscreens, they were originally designed for flat 2D screens and are neither ergonomic nor intuitive in 3D space. Physiological biometrics, such as iris scanning [5] and EEG [25], may require dedicated sensors to capture.

To address these limitations, researchers have begun exploring 3D-specific behavioral cues captured within the immersive environment for authentication, such as hand gestures [28] and eye-tracking [41], which better align with the natural interaction styles of VR/AR systems. However, the choice of action remains a critical factor: overly generic actions may lack uniqueness across users, while unfamiliar gestures may introduce unnecessary cognitive or physical burden.

In this paper, we introduce a new user interface for authentication in immersive VR/AR systems. This interface leverages behavioral biometrics, specifically in-air signing combined with signature recognition. In-air signatures have been explored for user identification [4, 27], signature verification [13] and authentication based on finger-motion sensing [24]. However, to the best of our knowledge, this approach has not yet been investigated for authentication in immersive VR/AR environments. Our gesture-based authentication interface allows users to authenticate by signing in the air, a behavior that is personal and identifiable.

### 1.1 Positioning Sign in the Air to Unlock as a user interface

One might argue that in-air signatures constitute merely a method or mode of input to VR/AR devices. However, we position our system as a new user interface modality rather than a simple input technique because it is designed to facilitate system entry and interaction in an integrated, user-centered manner. Unlike camera-based approaches, our interface relies solely on inertial sensing (specifically accelerometer data) for authentication. This sensing-based design enables users to perform signatures naturally while walking, turning, or facing in any direction, without needing to align themselves with a fixed viewpoint. In doing so, our system transforms the act of signing into an embodied, continuous interaction that integrates seamlessly into the immersive environment, fulfilling the criteria of a true interface rather than a peripheral input method. In contrast to traditional authentication mechanisms such as PINs or passwords, which require user disengagement and disrupt presence, our interface remains sensor-based and environment-integrated. Users authenticate through natural hand movements without secondary devices or external sensors, creating an intuitive and frictionless process that is both secure and personalized.

Table 1. Comparison of authentication interfaces in immersive environments.

| Criterion | **Traditional Interfaces** *(PIN, Passwords, Patterns)* | **Non-Traditional Interfaces** *(EEG, Eye-Tracking, Wearables)* | **Proposed User Interface** *(Gesture-based In-air Signature)* |
|---|---|---|---|
| **User Interaction** | Text/pattern input on 2D screens. | EEG headbands; gaze focus; paired wearables needed for use. | Sign/gesture in the air within VR/AR. |
| **Immersiveness** | External devices interrupt immersion. | Setting up Sensors/paring wearable break immersion. | Integrated in-headset; no extra hardware. |
| **Ergonomics** | Awkward in 3D because virtual keyboards and pointing gestures lack tactile feedback.. | EEG/gaze fatigue; wearables discomfort. | Fully embedded in the immersive session, no secondary device or sensor. |
| **Security** | Vulnerable to observation (shoulder surfing, replay). | Spoofing/interference; device leaks. | Unique behavioral biometrics; harder to forge. |
| **Ease of Use** | Requires memorization or precise input. | Calibration/sync needed. | relying on a familiar signing action. |
| **Privacy** | Credentials stored externally. | Sensitive physio data collected. | Local processing; minimal exposure. |
| **Adaptability in VR/AR** | Poor for immersive 3D. | Hard to scale in dynamic VR/AR. | Local, transient processing using only gesture trajectories is used |

Table 1 highlights how the Sign in the Air to Unlock interface overcomes limitations found in both traditional and emerging authentication methods. Conventional approaches such as PINs and pattern locks are inherently two-dimensional and interrupt the immersive flow by requiring users to disengage from the headset or use external devices [29]. Non-traditional interfaces such as EEG headbands, eye-tracking cameras, and wearable sensors like smartwatches, smart rings, or Electromyography armbands, introduce new challenges: EEG systems are intrusive and uncomfortable for extended wear [10], eye-tracking demands continuous visual focus and risks fatigue [12], and wearable sensors depend on pairing [31], calibration, or skin contact that can restrict movement [21] and raise privacy concerns [15]. In contrast, the in-air signature interface is entirely embedded within the immersive environment and relies on natural, familiar gesture (signature) that users already perform with their hands. It requires no secondary hardware, minimizes cognitive and physical load, and leverages behavioral biometrics that are both distinctive and privacy-preserving. By integrating authentication into the user's natural interaction flow, this interface provides a seamless, ergonomic, and secure alternative that aligns with the design principles of immersive human-computer interaction.

For evaluation part, we present PV-Net, a Point-Voxel Cross-Attention Network that can be used to authenticate users through their signatures signing in the air in immersive settings. PV-Net uses Inertial Measurement Units (IMUs) on controllers to capture 3D mid-air signature paths to check the identity of users. Signing is a common, natural, and very personal behavior, which makes it a great way to use biometric authentication in virtual spaces. PV-Net uses a dual-branch neural architecture to process both point-based and voxel-based representations of the input trajectory at the same time. The point branch gets spatial cues that are specific to a certain area, and the voxel branch uses a 3D convolutional pipeline to encode more general structural features. The multi-head cross-attention mechanism lets points and voxels pay attention to each other in real time, learning how to match up local and global representations in very small ways.

We evaluate PV-Net on two datasets: (1) the public **DeepAirSig** Malik et al. [27] dataset containing data collected from 40 participants through a GoPro camera, and (2) **ImmAirSig**, a self-collected dataset, containing data from 22 participants who signed in the air using Meta Quest 2 headset and paired controllers. The results show that PV-Net achieves strong performance on the DeepAirSig dataset with an EER of 2.5% (54.5% reduction in error rate relative to the baseline model.), and promising generalizability in headset-based environments with an accuracy of 76% on the the ImmAirSig.

This work makes the following key contributions:

- Investigate the challenges of adapting in-air signature authentication to fully immersive VR settings. Our evaluation shows that VR has some unique sources of performance variability, like inconsistent controller grip, user orientation drift, and spatial instability. Even though these things cause a moderate drop in accuracy, our findings provide empirical insight into the design considerations necessary for robust behavioral authentication in dynamic 3D environments. This analysis extends prior work by grounding performance evaluation in real-world immersive use cases.
- Introduce in-air signature authentication to immersive VR/AR for the first time. Our method addresses the challenge of seamless reauthentication within and across sessions. We present a new interface that combines signature recognition with dynamic gesture input, enabling effortless and ergonomically aligned interactions that feel natural to users in immersive spaces.
- Propose PV-Net, a dual-branch deep neural network for immersive authentication that fuses point-based and voxel-based representations of in-air signatures through a bidirectional cross-attention mechanism. This combination enables joint modeling of local motion dynamics and global spatial structure.
- Develop a cross-modal temporal learning design, combining Long Short-Term Memory(LSTM)-based sequential modeling with multi-head attention across point and voxel features, improving the network's ability to capture trajectory-level temporal context in 3D signature data.
- Evaluate PV-Net on two in-air signature datasets: (1) the public DeepAirSig dataset collected with depth camera, where our method achieves a 2.5% EER (a 54.5% reduction compared to the best baseline) and (2) a new dataset, **ImmAirSig**, which we collected using Meta Quest 2 in immersive VR environment, providing the first benchmark of signature verification performance in immersive environments.

## 2 PRIOR WORK

Authentication usability and security challenges are unique in immersive environments. Because unlike traditional platforms, VR/AR systems should give fully immersive experience to the user and absorb them into virtual or mixed environment. So making any interruption such as switching to a mobile phone or using a virtual keyboard breaks this flow.

Traditional authentication mechanisms, such as PINs and pattern locks, have often been adapted for a VR/AR environment to retain a feeling of familiarity. Yu et al. [40] provided a PIN system in VR using a virtual keyboard on which users could enter numbers through hand controllers. Likewise, Yadav et al. [39] implemented a pattern lock authentication system for head-mounted displays that emulates Android-style patterns via mid-air gestures. However, even though these approaches attempt to replicate successful mobile-based schemes, they often fall short in immersive contexts due to awkward 3D interaction, lack of tactile feedback, and susceptibility to shoulder surfing.

To overcome the limitations of knowledge-based and physiological authentication methods in VR/AR systems, researchers have started to focus on 3D behavioral cues such as in-air gestures. The idea was using built-in sensors such as IMUs, depth sensors, and stereo or Red, Green, Blue (RGB) cameras. Using these cues leading to more natural and immersive interaction without interrupting the user experience. Initial research was conducted on vision-based and handcrafted feature techniques for mid-air signature verification. Khoh et al. [20] used deep features extracted through transfer learning on the VGG-16 model to classify in-air hand gesture signatures captured by the Leap Motion device for both user verification and forgery detection. On the other hand, Guerra-Segura et al. [14] adopted a statistical approach, they modeled temporal features from 3D trajectories using Leap Motion sensor, and employed Least Squares Support Vector Machines (LS-SVMs) for authentication. However, their evaluation was based on a controlled and non-public dataset, which may limit the applicability of their results in dynamic environments. Malik et al. [26] created 3DAirSig, a multimodal in-air signature framework that uses synchronized multi-view video recordings and 3D trajectory data from a depth sensor. They made a depth-enriched in-air signature dataset by using CNN-based hand pose estimation to get fingertip trajectories. They also showed that depth alone can be a strong feature for verifying users. The system does need a lot of cameras, though. For example, it needs three GoPro cameras to record from different angles. This makes it hard to use in the real world and makes it hard to make it bigger.

Later work added other types of sensing to in-air authentication. Wazir et al. [38] proposed a doodle-based user authentication system using inertial data from wearable sensors. They designed a hybrid deep learning architecture constituted by CNNs and RNNs that exploit spatio-temporal dynamics with respect to user-drawn doodles. The system was very accurate, but we still don't know enough about how well it can handle sophisticated mimicry or imitation attacks in real-world situations. Guo and Sato [16] came up with a smartwatch-based in-air signature verification system that uses both curve-based and semantic-level features taken from wrist motion signals. Their method uses a dual-branch neural architecture to capture both fine-grained trajectory patterns and high-level writing semantics at the same time. The study did not test the system's long-term stability or repeatability, so it is not clear how well it would work over long periods of time. It was only useful for short-term user verification. Jung et al. [18] showed a way to identify users without having them touch anything. It uses Wi-Fi Channel State Information (CSI) signals made by handwritten signatures in the air. The system uses changes in CSI caused by the way a user writes to train a classifier for biometric identification. This shows how useful RF-based behavioral biometrics can be. However, the approach's sensitivity to noise in the environment and multipath interference makes it very hard to scale up and use in places where you can't control the environment.

New ways of interacting with immersive systems have led to new ways of authenticating. Wang et al. [37] have introduced a VR authentication scheme that is easy to remember. Users identify themselves by choosing moving groups of 3D objects that are set up in different ways in space. This design tries to find a middle ground between usability and security by using semantic and spatial cues to make things easier to remember. But user studies showed that the method works, though it might not work as well when people are stressed, have a lot on their minds, or are short on time. De Lorenzis et al. [8] proposed 3DK-Reate, as a Metaverse authentication system, it lets users build their own 3D keys using virtual blocks. The system uses asymmetric cryptography and a challenge–response method to verify users securely in social VR environments. However, it has not yet been tested in real-world conditions where network delays and performance issues could affect usability. Duezguen et al. [9] introduced ZeTA. It's a zero-trust authentication protocol that is designed for VR/AR head-mounted displays. It uses semantic challenge–response tasks, and it relies solely on built-in sensors like voice control, head movement, or touch input. The system is designed to resist shoulder-surfing attacks and support cross-cultural usability. However, it lacks empirical usability evaluation and may impose cognitive demands, especially under rapid or distracted usage conditions.

Central to this line of research, Malik et al. [27] introduced DeepAirSig, an end-to-end framework that uses autoencoder-based representations of in-air signatures collected via depth cameras. They achieved very good verification accuracy on a dataset of 1,800 samples from 40 users. However, their method relies on single-modality inputs and lacks temporal modeling, and the dataset was captured in a constrained, camera-based setting. In contrast, our work introduces PV-Net, a dual-branch architecture with point-voxel fusion and temporal attention. We test its performance not only on DeepAirSig but also in fully immersive VR, showing the problems and possibilities for behavioral authentication in real-world 3D environments.

## 3 PV-NET: A DUAL-BRANCH POINT-VOXEL CROSS-ATTENTION NETWORK

In VR/AR environments, users interact naturally within a three-dimensional (3D) space, making it difficult to constrain actions such as signing and drawing to a virtual two-dimensional (2D) plane. Without tactile feedback from a physical surface, accurately reproducing a signature on an invisible flat plane can be unintuitive and imprecise. Consequently, signatures captured in immersive settings using depth cameras or motion tracking sensors are recorded as sequences of $(x, y, z)$ coordinates, rather than traditional planar $(x, y)$ trajectories:

$$S_n = [(x_1, y_1, z_1), (x_2, y_2, z_2), ..., (x_n, y_n, z_n)]$$

The additional depth dimension enriches the spatial representation of signing behaviors, capturing not only lateral motion but also vertical variations, allowing the authentication system to incorporate spatial awareness, in-air motion dynamics, and the natural flow of hand movement.

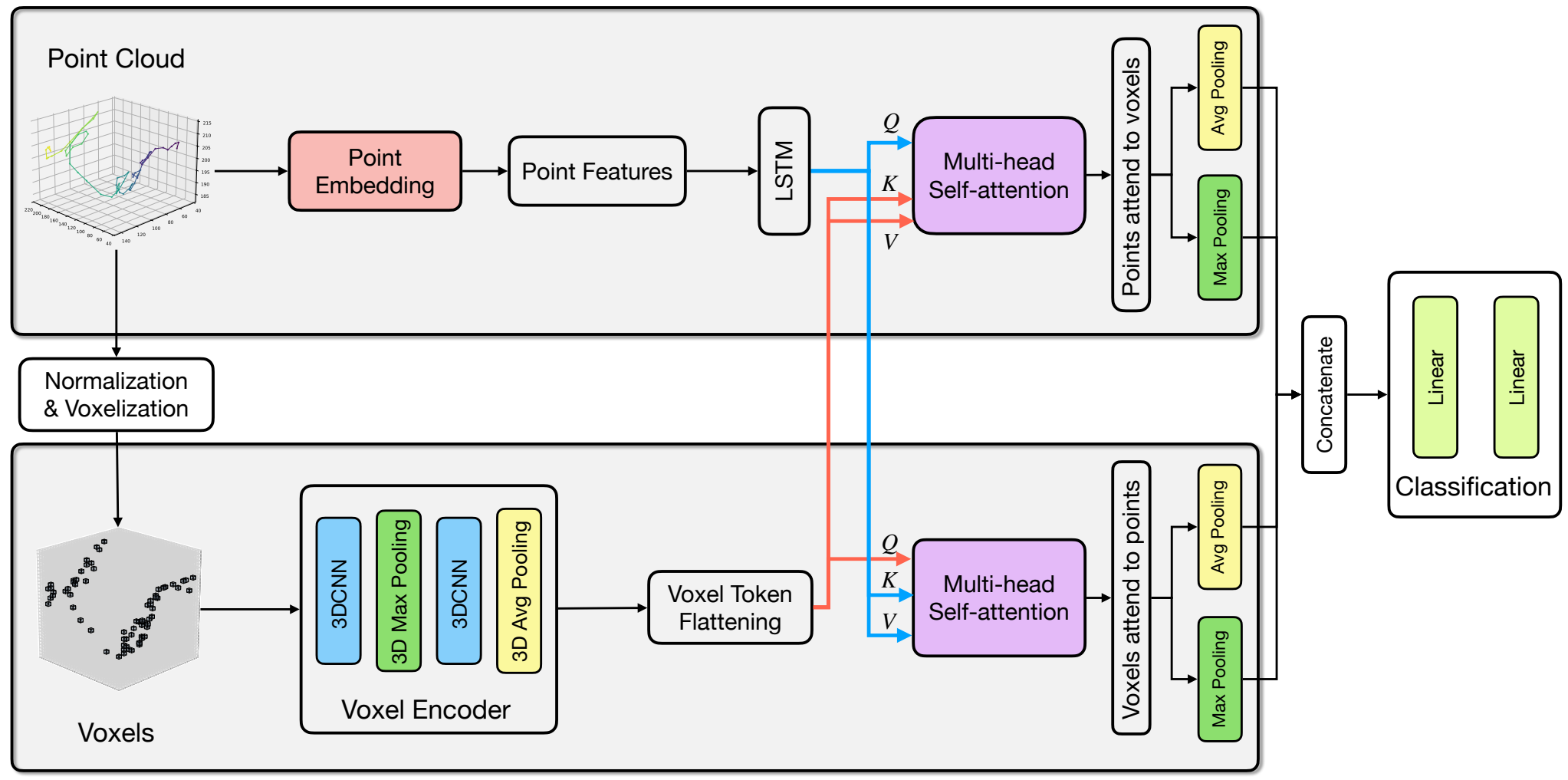

Fig. 1. Overview of the PV-Net architecture. The input signing trajectory in 3D space is processed by a dual-branch cross-attention mechanism. The point cloud branch captures fine-grained, point-wise details along the temporal signing sequence, while the voxel branch emphasizes the overall spatial structure of the signature, providing tolerance to intra-user variability over time.

### 3.1 Point cloud branch: capturing fine-grained, point-wise details

Earlier work by Malik et al. [27] has shown that treating the signing trajectory as a point cloud is effective for user authentication, as the overall spatial shape of a user's signature tends to remain consistent. However, while the signature trajectory is inherently sequential, point clouds treat each point as an independent and unordered entity, thereby ignoring temporal information such as signing speed and stroke acceleration. To incorporate both spatial and temporal aspects, we first model the 3D coordinate trajectory as a point cloud to capture its structural characteristics. We then introduce an LSTM layer that processes the point-wise features in their original signing order, enabling the model to also learn temporal dynamics. The upper branch in Figure 1 illustrates this point cloud processing stream.

Each captured trajectory, represented as a set of 3D points $S_n$, is first standardized to a fixed length $S_N$ to ensure consistency across samples. Trajectories longer than $N$ points are truncated, while shorter ones are zero-padded at the end. The resulting sequence is then treated as the input point cloud for the point cloud branch.

A point embedding module, consisting of a linear layer followed by layer normalization, ReLU activation, and dropout, is applied to each point to project it into a higher-dimensional feature space, resulting in a sequence of point features $PF \in \mathrm{R}^{N\times d}$. Equation 1 describes the point cloud feature extraction module.

$$p' = \mathrm{ReLU}(\mathrm{Dropout}(\mathrm{LayerNorm}(\mathrm{Linear}(A \cdot p/D)))) \tag{1}$$

where $p$ and $p'$ denote the input and output point feature matrices at a given layer, respectively.

To incorporate temporal context from the signing process, we introduce a unidirectional LSTM layer that processes the point features $PF$ in their original signing order. This allows the model to capture sequential dependencies such as stroke direction, velocity, and acceleration not captured by the point cloud representation alone. The resulting temporally enriched features are then passed to the cross-attention module, where they interact with the voxel branch to integrate local motion details with global shape information.

### 3.2 Voxel branch: capturing global structural patterns

Although the point cloud branch captures both spatial and temporal details of the signing trajectory, its focus on fine-grained, point-level features may lead the model to overlook the global structural patterns of the signature, such as its overall shape and spatial distribution. In real-world in-air signing scenarios, the absence of a tangible surface makes it even more difficult for users to consistently reproduce their signatures over time compared to signing on a 2D surface. This results in greater intra-user variations, which pose a critical challenge that authentication models must address.

To complement the localized perspective of the point cloud branch, we introduce the voxel branch, which is designed to learn a holistic representation of the signature by capturing its global spatial configuration. As depicted in the lower branch of Figure 1, the input point cloud is first normalized by translating and scaling its bounding box to fit within a unit cube. It is then voxelized [17] by dividing the normalized 3D space into a uniform grid of size $32 \times 32 \times 32$. Each voxel is assigned a value based on the number of points it contains, followed by a logarithmic scaling to reduce the dynamic range and stabilize training. This voxelized representation transforms the irregular and sparse point cloud into a dense volumetric grid, which suppresses small-scale spatial variations caused by rapid or hesitant signing motions while preserving the global spatial occupancy patterns of the signature.

A voxel feature encoder is then applied, consisting of two blocks of 3D convolution, batch normalization, and ReLU activation, with a max-pooling layer following the first block and an adaptive average pooling layer at the end. This encoder extracts global spatial features across multiple scales, enabling the model to capture the overall structure and shape contours of the signature trajectory, and enhancing its robustness to local inconsistencies induced by variations in signing speed or minor movement fluctuations.

Finally, the output feature map from the voxel encoder is flattened along the spatial dimensions to form a sequence of voxel tokens $V$, which are passed to the cross-attention module to interact with features from the point cloud branch.

### 3.3 Cross-attention module: modeling interactions between point and voxel features

After extracting localized features from the point cloud branch and global structural features from the voxel branch, we leverage a cross-attention module by applying two multi-head attention layers to model the interactions between these two complementary representations, as shown in Figure 1.

In the first Point-to-Voxel (P2V) attention, represented in equation 2, the point features $P$ attend to the voxel features $V$ by using $P$ as queries and $V$ as keys and values in the attention mechanism. This allows each point feature to dynamically incorporate broader spatial context from the global voxel representation. In the second Voxel-to-Point (V2P) attention, as shown in equation 3, the voxel features $V$ attend to the point features $P$ by using $V$ as queries and $P$ as keys and values. This enables each voxel feature to selectively aggregate fine-grained local information from the point cloud.

$$\text{P2V} = \text{MultiHeadAttention}(\text{query} = P,\ \text{key} = V,\ \text{value} = V) \tag{2}$$

$$\text{V2P} = \text{MultiHeadAttention}(\text{query} = V,\ \text{key} = P,\ \text{value} = P) \tag{3}$$

The outputs from the V2P and P2V attentions are then processed separately using both mean pooling and max pooling across the token sequences. The resulting pooled representations are concatenated to form the final feature vector, which is used for user authentication.

## 4 EXPERIMENTS AND EVALUATION ON CAMERA-BASED DATASET

### 4.1 Dataset description and pre-processing

To benchmark the performance of PV-Net against existing methods, we also evaluated it on the DeepAirSig dataset, a publicly available in-air signature dataset collected by Malik et al. [27]. In their work, three GoPro cameras were used to record hand movements in 3D space from different viewpoints while subjects signed their signatures in the air using their pointing finger. A CNN-based model was then trained to segment the hand region and extract fingertip movements as 3D coordinates representing the in-air signature trajectories. The dataset contains 20 signature trajectories for each of 40 subjects, with an even split into training and testing sets, each containing 10 samples per subject.

After data cleaning and pre-processing, data from 5 subjects were excluded due to missing values or incomplete recordings. The identification system is thus constructed using the remaining 35 subjects. For each subject, their training set samples are further divided into an 8:2 ratio for training and validation, while the original 10 testing samples are used exclusively for testing.

### 4.2 Experiment setup and training strategy

We make all signature trajectories in the same length by setting $N = 300$ as most signatures consist of around 200 coordinates.

The model is trained end-to-end using cross-entropy loss with label smoothing ( $\alpha = 0.1$ ) to reduce overconfidence in predictions. Optimization is performed using AdamW with a learning rate of $3 \times 10^{-4}$, weight decay $\left(10^{-4}\right)$, and gradient clipping (max norm = 5 ). The learning rate is dynamically adjusted via ReduceLROnPlateau, monitoring validation loss with a patience of 20 epochs. Early stopping halts training if validation accuracy does not improve for 30 epochs.

### 4.3 Experimental results and insights

We evaluate our proposed PV-Net model under three configurations: point cloud-only, voxel-only, and a fused representation combining both. Table 2 shows how the Equal Error Rates (EER) compare across the different settings and Figures 2-4 shows the ROC, score distribution, and FAR/FRR analysis for each setting. Among them, the full PV-Net model (Figure 4), which leverages both point and voxel features through a unified transformer architecture, achieves the lowest EER of 0.025, outperforming both the individual variants with 0.175 for point cloud-only and 0.150 for voxel-only.

Table 2. Performance comparison of different PV-Net input variants on the DeepAirSig dataset.

| Model Variant | AUC | EER |
|---|---|---|
| Point Cloud-Only | 0.90 | 0.175 |
| Voxel-Only | 0.95 | 0.150 |
| PV-Net | **1.00** | **0.025** |

To benchmark our method against prior work, we compared PV-Net with four baseline models reported by Malik et al. [27]. Table 3 shows that PV-Net has an Equal Error Rate (EER) of 2.5%, which is a significantly better than the best baseline, PointCloud-based verification, which had an EER of 5.5%. This result demonstrates PV-Net's strong performance on the DeepAirSig dataset using a cross-modal point-voxel representation.

Table 3. Comparison of Equal Error Rate (EER%) between PV-Net and baseline models from Malik et al. [27].

| Model | EER (%) |
|---|---|
| DTW 3D (X,Y,Z) | 17.0 |
| Deep One-Class SVDD | 20.7 |
| Image-based Verification | 26.0 |
| PointCloud-based Verification | 5.5 |
| **PV-Net (Ours)** | **2.5** |

As shown in Figure 4, the ROC curve for the full PV-Net model rises sharply toward the upper-left corner, indicating consistently high true-positive rates across a wide range of decision thresholds compared with the single-branches (Figures 2 and 3). The similarity-score distributions in the second columns reveal substantial overlap between genuine and forged samples. As shown in the full PV-Net model (Figure 4) there is a clear separation between the two classes compared with point cloud-only (Figure 2) and voxel-only models (Figure 3). This distinction indicates that the fused representation effectively discriminates between genuine and forged signatures. The intersection of the FAR and FRR curves at an EER of 2.5% demonstrates that our model achieves high performance between false acceptance and false rejection rates.

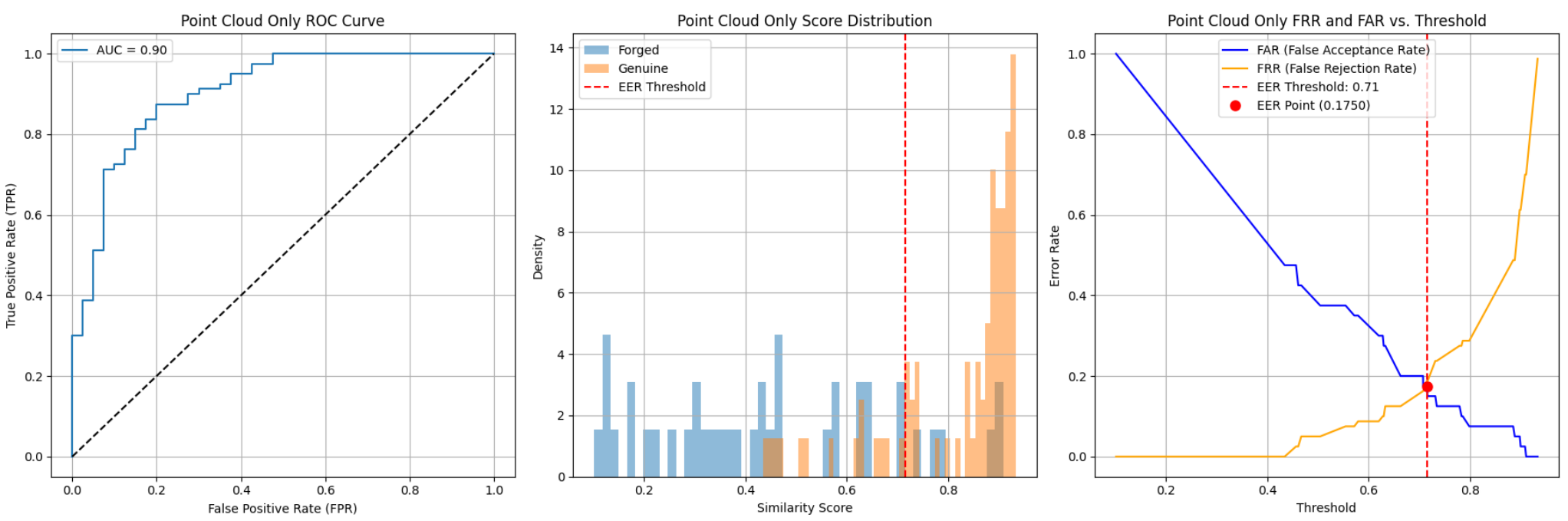

Fig. 2. Performance curves for the point cloud–only variant of PV-Net on DeepAirSig.

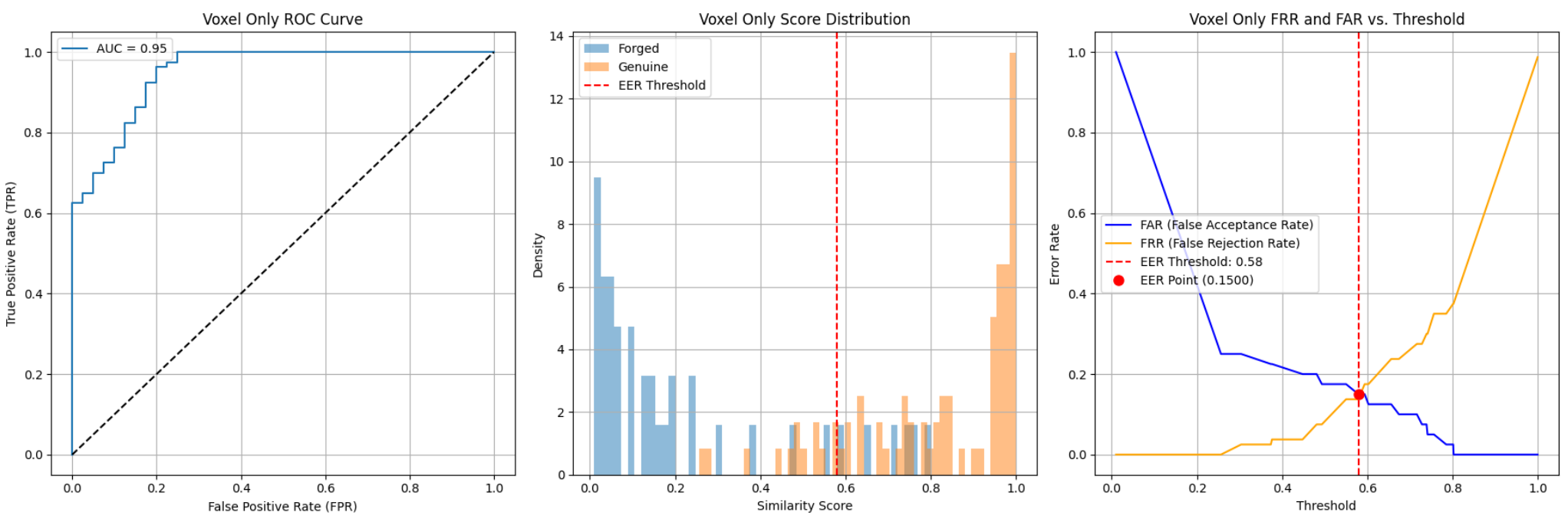

Fig. 3. Performance curves for the voxel-only variant of PV-Net on DeepAirSig.

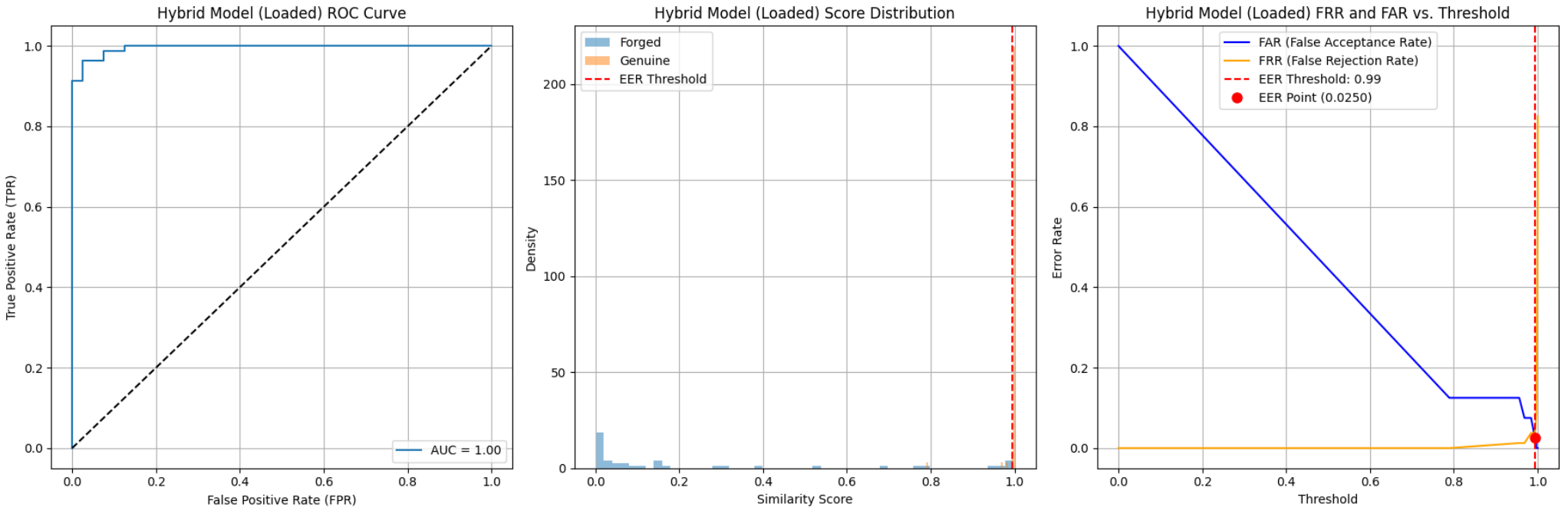

Fig. 4. Performance curves for the full PV-Net model on DeepAirSig (EER = 2.5%).

## 5 EVALUATION ON IMMAIRSIG

### 5.1 ImmAirSig Data collection and description

We used the Meta Quest 2 headset and controllers to collect a new in-air signature dataset for evaluating PV-Net in real-world immersive settings. Data collection was conducted in Gravity Sketch, a VR-based 3D design application that records natural hand trajectories as users interact freely in virtual space. Using handheld controllers, participants wrote signatures in mid-air. This process created a realistic simulation of biometric input within an immersive environment.

The dataset included 22 people, 16 men and 5 women, aged 22 to 73. One person didn't want to say what gender they were. Each participant provided 20 genuine signatures written in the air with their dominant hand, without constraints on orientation or writing direction (Figure 5). Then they labeled and stored them in the cloud environment, and later exported for evaluation. They were then asked to produce 20 forgeries, attempting to imitate the genuine signatures of randomly assigned participants without any boundaries in terms of keeping the same set up of genuine ones . Before trying to forge, participants could open the real sample in the Gravity sketch's environment using the cloud menu, walk around it in 3D environment and practice the genuine sample. This was meant to mimic real-life authentication situations where people try to copy something.

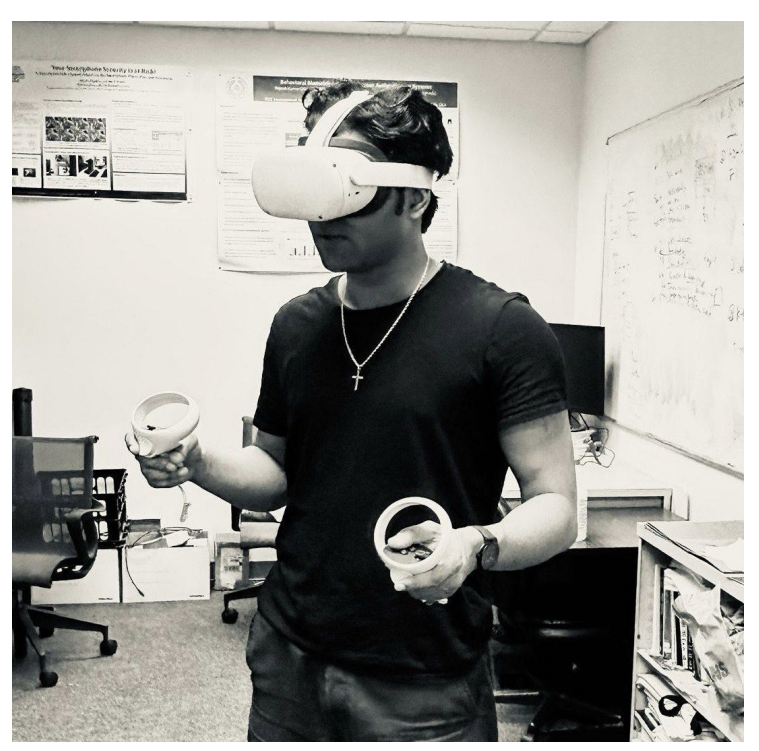

(a) Data collection process

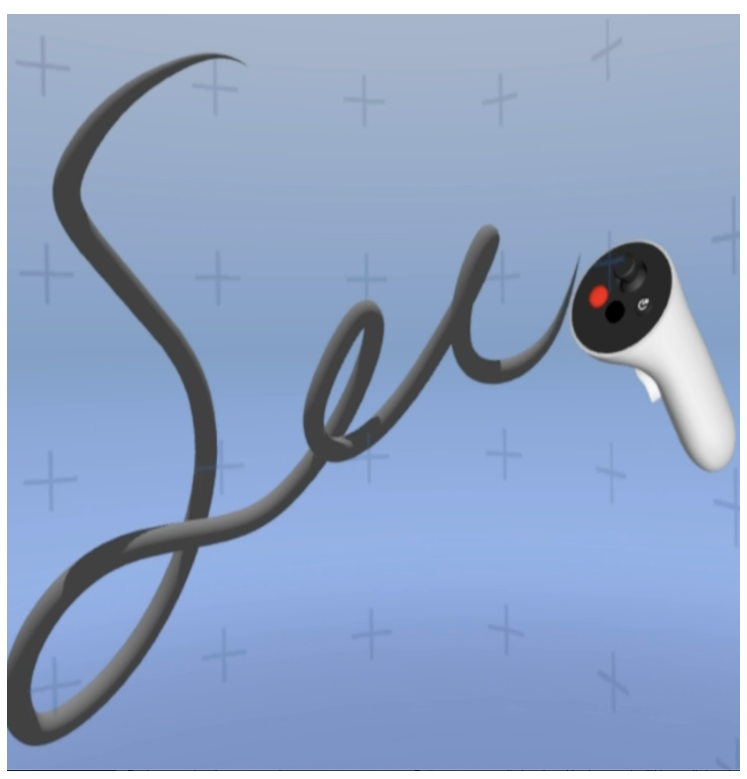

(b) Signing in VR with controllers

Fig. 5. Data collection and signing workflow in the immersive VR environment.

Figure 6 shows two 3D slots of in-air signature samples corresponding to the genuine (Figure 6a) and forged (Figure 6b) cases. The X, Y, and Z axes are color-coded to illustrate the unconstrained spatial environment in which the data were collected and it shows signatures are not anchored to a fixed position in the air and can vary freely across three dimensions. The dynamic motion of these samples are provided in the supplementary materials as short videos.

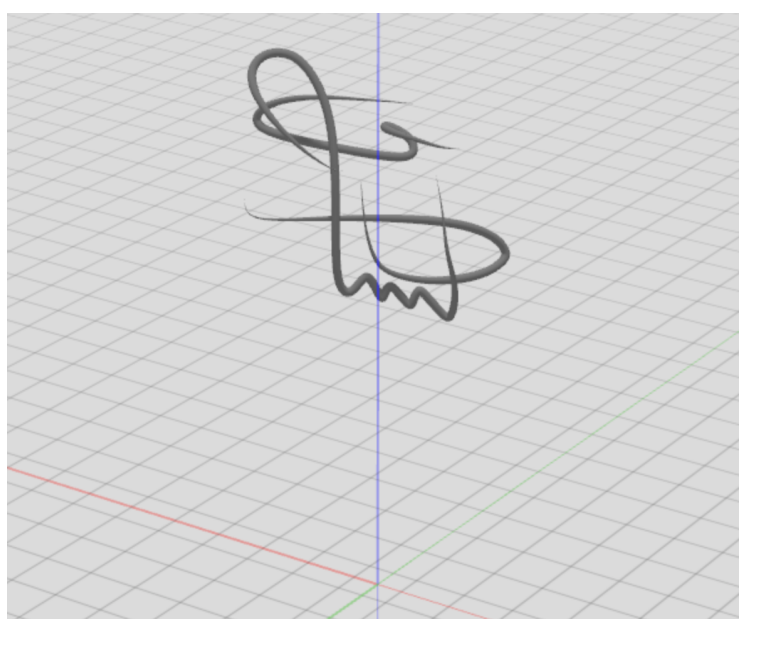

(a) Genuine

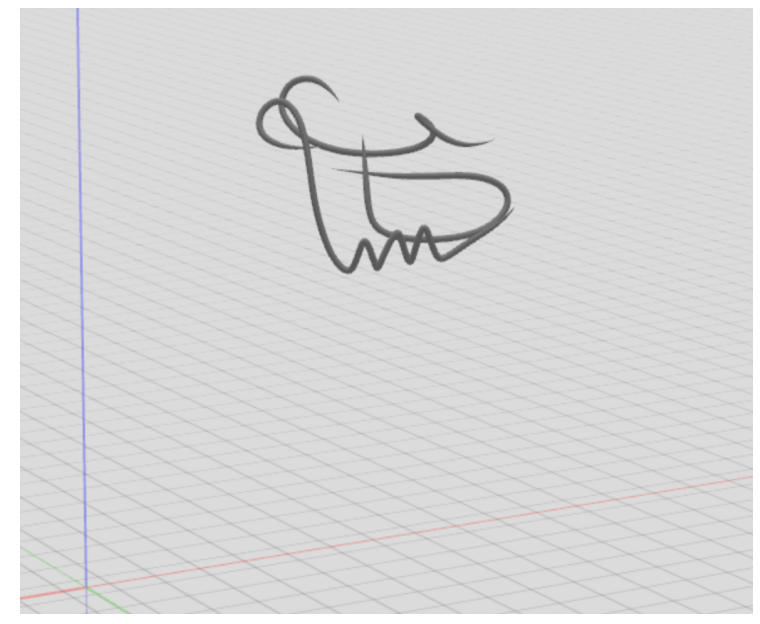

(b) Forged

Fig. 6. Samples of genuine and forged in-air signatures.

### 5.2 ImmAirSig Data pre-processing

The controller's motion path in the virtual space was recorded as a series of 3D coordinates (x, y, z) for each signature as an object file. We collected 880 samples including 440 genuine and 440 forged. To prepare the ImmAirSig data for model input, we processed raw point cloud recordings captured from the Meta Quest 2 headset. Each signature instance originally contained between 5,000 and 20,000 points stored in .obj format. We used a sliding window approach followed by temporal averaging within each window to make the sequence length the same for all users and cut down on noise. This process compresses the spatial trajectory while preserving the global signature structure. As a result, each signature was resampled to 6,000 points, which made sure that the input dimensions were the same across the whole dataset.

### 5.3 ImmAirSig experimental results and insights

We evaluated our model under the same three configurations: point cloud-only, voxel-only, and a fused representation combining both on VR dataset.

Table 4 shows how the EER compare across the different settings. The full PV-Net model, again shows the lowest EER of 0.242 outperforming the other configurations.

Table 4. Performance comparison of different PV-Net input variants on the VR dataset.

| **Model Variant** | **AUC** | **EER** |
|---|---|---|
| Point Cloud-Only | 0.6150 | 0.3895 |
| Voxel-Only | 0.6558 | 0.3684 |
| PV-Net | **0.8157** | **0.2421** |

The relatively higher EER of 24.2% in our VR-based authentication setup can be attributed, in part, to the stark contrast in data acquisition environments compared to the public DeepAirSig dataset. DeepAirSig used a very controlled capture environment that included Intel's Senz3D depth camera and positional markers to help users move their hands. Moreover, the three synchronized GoPro cameras made it possible to accurately estimate the trajectory of hand movement in 3D from multiple angles. Index fingertip tracking and consistent frontal recording angles made sure that the data collected was reliable and could be repeated.

On the other hand, our VR setup shows realistic and free user interactions, where signatures are captured with handheld controllers in immersive environments without touching the surface, getting spatial guidance, or using external markers. These differences introduce several compounding sources of noise: drift from the IMU sensors, varied user hand poses, inconsistent gesture scale, and the absence of proprioceptive anchoring. Additionally, VR changes the user's orientation and environment in real time, which makes model learning even harder than the depth camera setup's fixed viewpoint and lighting conditions.

Thus, while DeepAirSig benefits from multi-view video assistance and spatial anchoring, our VR-based system represents a more challenging, but more realistic use case that puts practical deployability ahead of laboratory precision. Our results demonstrate that PV-Net still generalizes reasonably well to this setting, which supports its potential for future in real-world biometric authentication scenarios.

As shown in Figures 7–9, the ROC, FAR/FRR, and score-distribution analyses illustrate how performance varies across PV-Net configurations in the immersive VR environment. The results show a clear trend of degradation as the capture conditions become less constrained. The full PV-Net model (Figure 9) still maintains a higher area under the ROC curve (0.82) compared with the single-branch variants, indicating that multimodal fusion continues to enhance discrimination even in the presence of sensor noise. In contrast to the controlled, camera-based desktop setting (DeepAirSig) results, (Figures 2–4), the similarity-score distributions obtained from the VR environment are more uniformly spread across a wider range. For both the point cloud-only (Figure 7) and voxel-only (Figure 8) variants, the absence of either temporal or spatial cues makes it difficult to identify a stable decision threshold, leading to greater overlap between genuine and forged samples. The full PV-Net model (Figure 9), however, shows a substantially smaller overlap, confirming that the fused representation effectively separates genuine from forged trajectories even under unconstrained VR capture condit

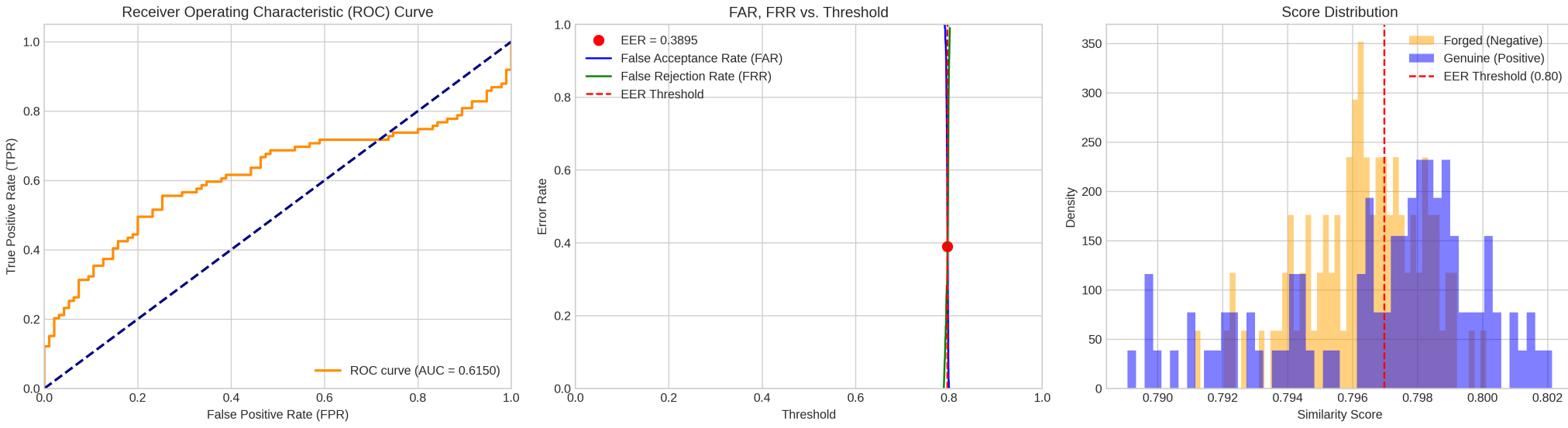

Fig. 7. Performance curves for the point cloud–only variant of PV-Net on ImmAirSig.

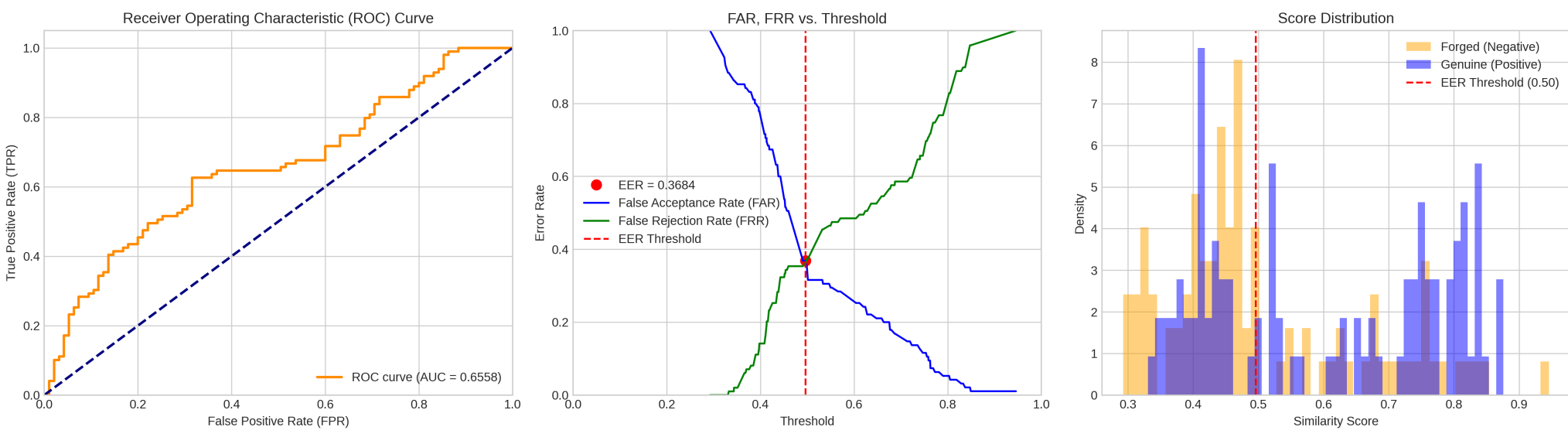

Fig. 8. Performance curves for the voxel–only variant of PV-Net on ImmAirSig.

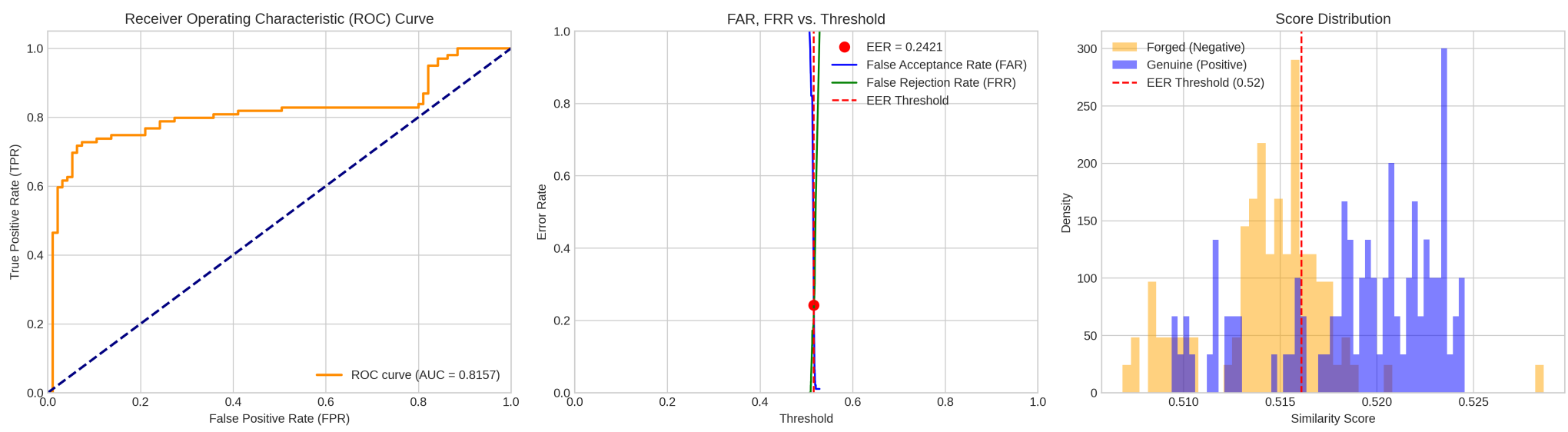

Fig. 9. Performance curves for the full PV-Net model on ImmAirSig. The model achieves an EER of 24.21%, demonstrating the benefit of multimodal fusion.

## 6 CONCLUSION

This work, Sign in the Air to Unlock, is a new authentication interface for immersive VR environments, powered by PV-Net, a dual-branch network that fuses point- and voxel-based representations through cross-attention. By transforming in-air signing into an embodied interaction, our approach enables secure and natural authentication without disrupting immersion. Through experiments on both the public DeepAirSig dataset and the newly collected ImmAirSig dataset, we demonstrated the robustness and adaptability of PV-Net across sensing conditions. PV-Net achieved an EER of 2.5% on DeepAirSig that represents a 54.5% improvement over the best prior baseline and maintained a promising 76% classification accuracy on ImmAirSig, despite the inherent variability and sensor drift of consumer-grade VR devices. These findings validate the feasibility of integrating behavioral authentication directly into embodied VR interaction, without interrupting immersion or requiring external hardware. Future work will involve improving signature segmentation methods, integrating motion context, and exploring user adaptation strategies for improved generalizability.